\documentclass[final]{cvpr}

\usepackage{times}
\usepackage{epsfig}
\usepackage{graphicx}
\usepackage{amsmath}
\usepackage{amssymb}

\usepackage{algorithm}
\usepackage{algorithmic}
\usepackage{bm}
\usepackage{mathtools}
\usepackage{makecell}
\usepackage{enumitem}
\usepackage{caption}
\usepackage{lipsum}
\usepackage{cuted}
\usepackage{siunitx}
\usepackage{booktabs}

\usepackage{subcaption}
\usepackage{multirow}
\makeatletter
\DeclareRobustCommand\onedot{\futurelet\@let@token\@onedot}
\def\@onedot{\ifx\@let@token.\else.\null\fi\xspace}

\newcommand{\vect}[1]{\boldsymbol{\mathbf{#1}}}
\def\eg{\emph{e.g}\onedot} 
\def\ie{\emph{i.e}\onedot} 
 
 \def\vs{\emph{vs}\onedot}

\makeatother

\usepackage[pagebackref=true,breaklinks=true,colorlinks,bookmarks=false]{hyperref}

\def\cvprPaperID{7140} 
\def\confYear{CVPR 2021}

\begin{document}

\title{RepVGG: Making VGG-style ConvNets Great Again}

\author{Xiaohan Ding \textsuperscript{1}\thanks{This work is supported by The National Key Research and Development Program of China (No. 2017YFA0700800), the National Natural Science Foundation of China (No.61925107, No.U1936202) and Beijing Academy of Artificial Intelligence (BAAI). Xiaohan Ding is funded by the Baidu Scholarship Program 2019. This work is done during Xiaohan Ding and Ningning Ma's internship at MEGVII Technology.} \quad Xiangyu Zhang \textsuperscript{2} \quad Ningning Ma \textsuperscript{3} \\
	Jungong Han \textsuperscript{4} \quad Guiguang Ding \textsuperscript{1}\thanks{Corresponding author.} \quad Jian Sun \textsuperscript{2} \\
	\textsuperscript{1} Beijing National Research Center for Information Science and Technology (BNRist); \\School of Software, Tsinghua University, Beijing, China \\
	\textsuperscript{2} MEGVII Technology \\
	\textsuperscript{3} Hong Kong University of Science and Technology \\
	\textsuperscript{4} Computer Science Department, Aberystwyth University, SY23 3FL, UK \\
	\tt\small dxh17@mails.tsinghua.edu.cn \quad zhangxiangyu@megvii.com \quad nmaac@cse.ust.hk\\
	\tt\small jungonghan77@gmail.com \quad dinggg@tsinghua.edu.cn \quad sunjian@megvii.com \\
}

\maketitle
\pagestyle{empty}
\thispagestyle{empty}

\begin{abstract}
	We present a simple but powerful architecture of convolutional neural network, which has a VGG-like inference-time body composed of nothing but a stack of $3\times3$ convolution and ReLU, while the training-time model has a multi-branch topology. Such decoupling of the training-time and inference-time architecture is realized by a structural re-parameterization technique so that the model is named RepVGG. On ImageNet, RepVGG reaches over 80\% top-1 accuracy, which is the first time for a plain model, to the best of our knowledge. On NVIDIA 1080Ti GPU, RepVGG models run 83\% faster than ResNet-50 or 101\% faster than ResNet-101 with higher accuracy and show favorable accuracy-speed trade-off compared to the state-of-the-art models like EfficientNet and RegNet. The code and trained models are available at \url{https://github.com/megvii-model/RepVGG}.
\end{abstract}

\section{Introduction}

A classic Convolutional Neural Network (ConvNet), VGG \cite{simonyan2014very}, achieved huge success in image recognition with a simple architecture composed of a stack of conv, ReLU, and pooling. With Inception \cite{szegedy2015going,szegedy2016rethinking,szegedy2017inception,ioffe2015batch}, ResNet \cite{he2016deep} and DenseNet \cite{huang2017densely}, a lot of research interests were shifted to well-designed architectures, making the models more and more complicated. Some recent architectures are based on automatic \cite{zoph2018learning,real2019regularized,liu2018progressive} or manual \cite{regnet} architecture search, or a searched compound scaling strategy~\cite{efficientnet}.

\begin{figure}
	\begin{subfigure}{0.49\linewidth}
		\includegraphics[width=\linewidth]{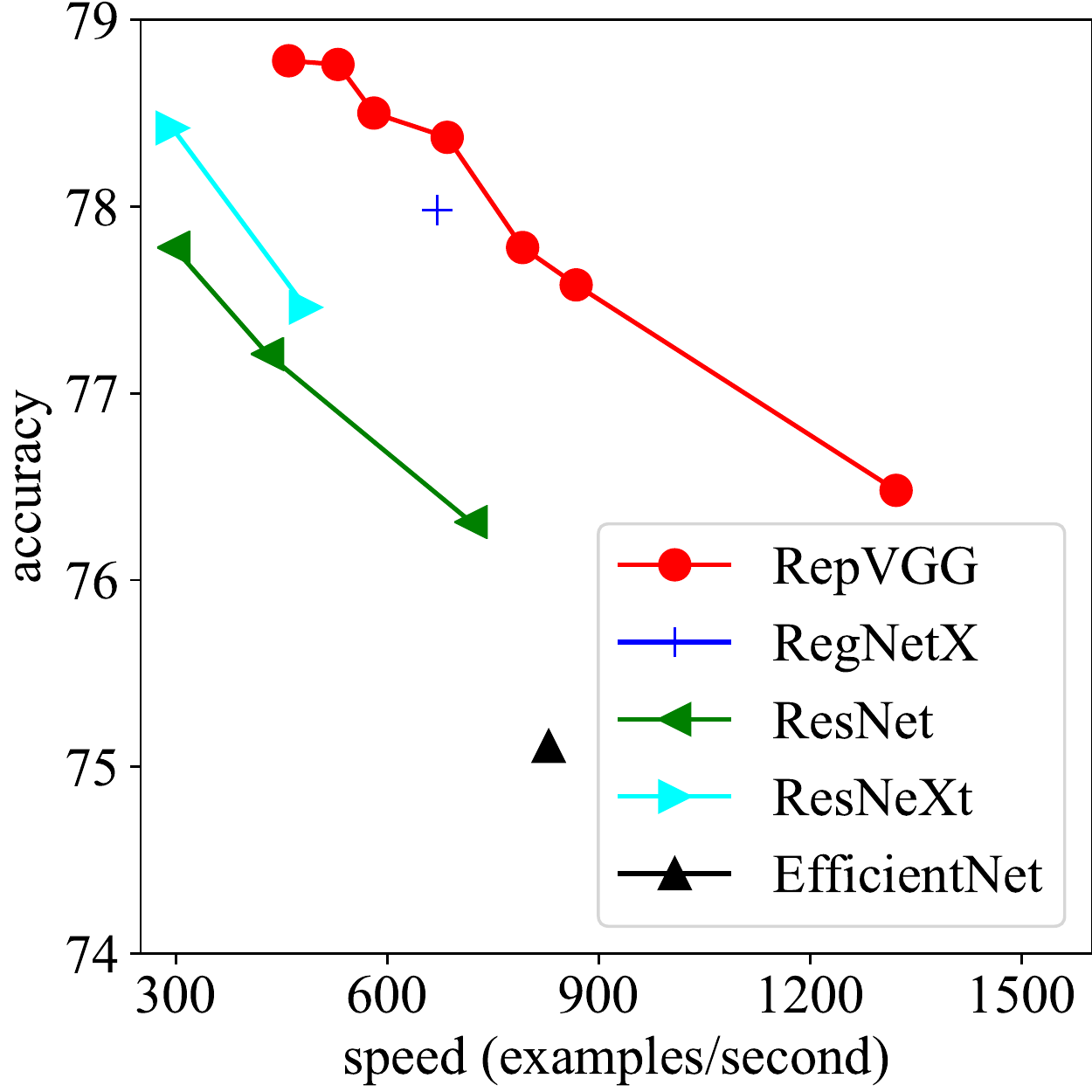}
		\label{fig-tradeoff-weak} 
	\end{subfigure}
	\begin{subfigure}{0.49\linewidth}
		\includegraphics[width=\linewidth]{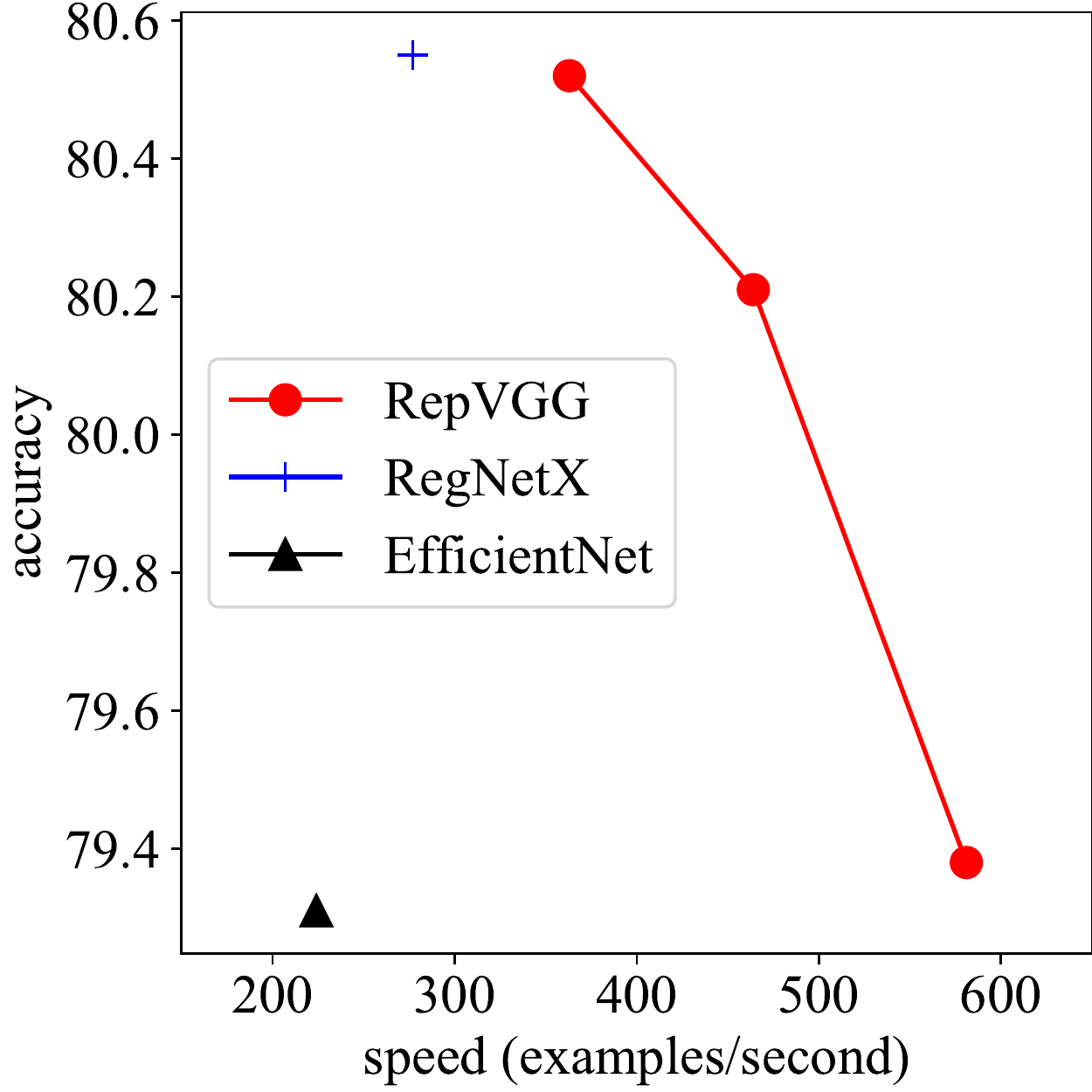}
		\label{fig-tradeoff-strong} 
	\end{subfigure}
	\vskip -0.2in
	\caption{Top-1 accuracy on ImageNet \vs actual speed. Left: lightweight and middleweight RepVGG and baselines trained in 120 epochs. Right: heavyweight models trained in 200 epochs. The speed is tested on the same 1080Ti with a batch size of 128, full precision (fp32), single crop, and measured in examples/second. The input resolution is 300 for EfficientNet-B3 \cite{efficientnet} and 224 for the others.}
	\label{fig-tradeoff}
	\vskip -0.15in
\end{figure}
\begin{figure}
	\begin{center}
		\includegraphics[width=0.85\linewidth]{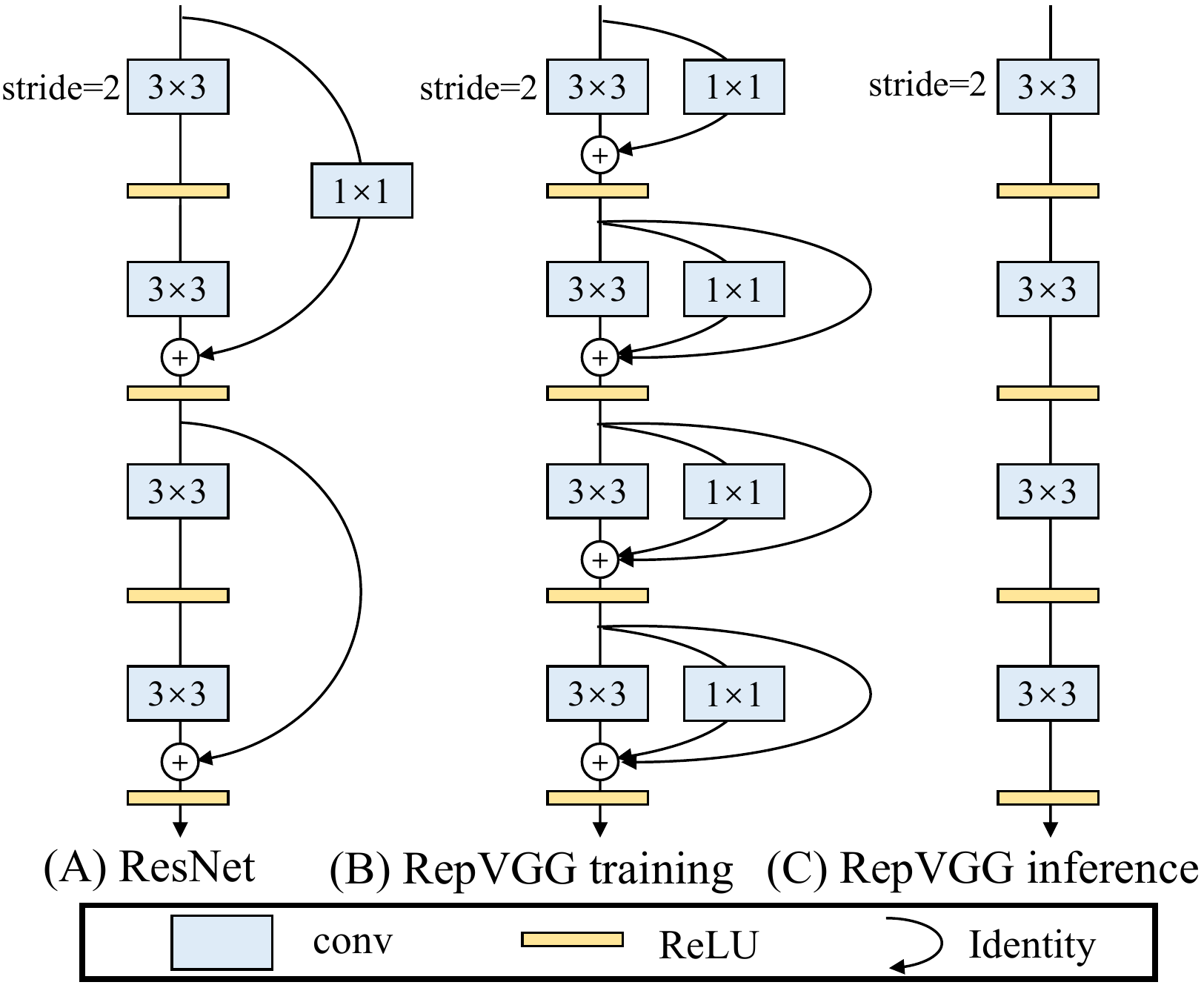}
		\vspace{-0.10in}
		\caption{Sketch of RepVGG architecture. RepVGG has 5 stages and conducts down-sampling via stride-2 convolution at the beginning of a stage. Here we only show the first 4 layers of a specific stage. As inspired by ResNet \cite{he2016deep}, we also use identity and $1\times1$ branches, but only for training.}
		\label{fig-arch}
		\vspace{-0.2in}
	\end{center}
\end{figure}

Though many complicated ConvNets deliver higher accuracy than the simple ones, the drawbacks are significant. \textbf{1)} The complicated multi-branch designs (\eg, residual-addition in ResNet and branch-concatenation in Inception) make the model difficult to implement and customize, slow down the inference and reduce the memory utilization. \textbf{2)} Some components (\eg, depthwise conv in Xception \cite{chollet2017xception} and MobileNets \cite{mbv1,mbv2} and channel shuffle in ShuffleNets \cite{ma2018shufflenet,zhang2018shufflenet}) increase the memory access cost and lack supports of various devices. With so many factors affecting the inference speed, the amount of floating-point operations (FLOPs) does not precisely reflect the actual speed. Though some novel models have lower FLOPs than the old-fashioned ones like VGG and ResNet-18/34/50~\cite{he2016deep}, they may not run faster (Table. \ref{table-weak}). Consequently, VGG and the original versions of ResNets are still heavily used for real-world applications in both academia and industry.

In this paper, we propose RepVGG, a VGG-style architecture which outperforms many complicated models (Fig. \ref{fig-tradeoff}). RepVGG has the following advantages.
\begin{itemize}[noitemsep,nolistsep,topsep=0pt,parsep=0pt,partopsep=0pt]
	\item The model has a VGG-like plain (a.k.a. feed-forward) topology \footnote{In this paper, a network \textit{topology} only focuses on how the components connect to others, an \textit{architecture} refers to the topology together with the specification of components like depth and width, and a \textit{structure} may refer to any component or part of the architecture.} without any branches, which means every layer takes the output of its only preceding layer as input and feeds the output into its only following layer.
	\item The model's body uses only $3\times3$ conv and ReLU.
	\item The concrete architecture (including the specific depth and layer widths) is instantiated with no automatic search \cite{zoph2018learning}, manual refinement \cite{regnet}, compound scaling \cite{efficientnet}, nor other heavy designs.
\end{itemize}

It is challenging for a plain model to reach a comparable level of performance as the multi-branch architectures. An explanation is that a multi-branch topology, \eg, ResNet, makes the model an implicit ensemble of numerous shallower models~\cite{veit2016residual}, so that training a multi-branch model avoids the gradient vanishing problem.

Since the benefits of multi-branch architecture are all for training and the drawbacks are undesired for inference, we propose to \textit{decouple the training-time multi-branch and inference-time plain architecture} via \textit{structural re-parameterization}, which means converting the architecture from one to another via transforming its parameters. To be specific, a network structure is coupled with a set of parameters, \eg, a conv layer is represented by a 4th-order kernel tensor. If the parameters of a certain structure can be converted into another set of parameters coupled by another structure, we can equivalently replace the former with the latter, so that the overall network architecture is changed.

Specifically, we construct the training-time RepVGG using identity and $1\times1$ branches, which is inspired by ResNet but in a different way that the branches can be removed by structural re-parameterization (Fig. \ref{fig-arch},\ref{fig-transform}). After training, we perform the transformation with simple algebra, as an identity branch can be regarded as a degraded $1\times1$ conv, and the latter can be further regarded as a degraded $3\times3$ conv, so that we can construct a single $3\times3$ kernel with the trained parameters of the original $3\times3$ kernel, identity and $1\times1$ branches and batch normalization (BN) \cite{ioffe2015batch} layers. Consequently, the transformed model has a stack of $3\times3$ conv layers, which is saved for test and deployment. 

Notably, the body of an inference-time RepVGG only has one single type of operator: $3\times3$ conv followed by ReLU, which makes RepVGG fast on generic computing devices like GPUs. Even better, RepVGG allows for specialized hardware to achieve even higher speed because given the chip size and power consumption, the fewer types of operators we require, the more computing units we can integrate onto the chip. Consequently, an inference chip specialized for RepVGG can have an enormous number of $3\times3$-ReLU units and fewer memory units (because the plain topology is memory-economical, as shown in Fig. \ref{fig-memory}). Our contributions are summarized as follows.
\begin{itemize}[noitemsep,nolistsep,topsep=0pt,parsep=0pt,partopsep=0pt]
	\item We propose RepVGG, a simple architecture with favorable speed-accuracy trade-off compared to the state-of-the-arts.
	\item We propose to use structural re-parameterization to decouple a training-time multi-branch topology with an inference-time plain architecture.
	\item We show the effectiveness of RepVGG in image classification and semantic segmentation, and the efficiency and ease of implementation.
\end{itemize}

\section{Related Work}

\subsection{From Single-path to Multi-branch}

After VGG \cite{simonyan2014very} raised the top-1 accuracy of ImageNet classification to above 70\%, there have been many innovations in making ConvNets complicated for high performance, \eg, the contemporary GoogLeNet \cite{szegedy2015going} and later Inception models \cite{szegedy2016rethinking,szegedy2017inception,ioffe2015batch} adopted elaborately designed multi-branch architectures, ResNet \cite{he2016deep} proposed a simplified two-branch architecture, and DenseNet \cite{huang2017densely} made the topology more complicated by connecting lower-level layers with numerous higher-level ones. Neural architecture search (NAS) \cite{zoph2018learning,real2019regularized,liu2018progressive,efficientnet} and manual designing space design \cite{regnet} can generate ConvNets with higher performance but at the costs of vast computing resources or manpower. Some large versions of NAS-generated models are even not trainable on ordinary GPUs, hence limiting the applications. Except for the inconvenience of implementation, the complicated models may reduce the degree of parallelism \cite{ma2018shufflenet} hence slow down the inference.

\subsection{Effective Training of Single-path Models}\label{sect-related-work-2}

There have been some attempts to train ConvNets without branches. However, the prior works mainly sought to make the very deep models converge with reasonable accuracy, but not achieve better performance than the complicated models. Consequently, the methods and resultant models were neither simple nor practical. An initialization method~\cite{xiao2018dynamical} was proposed to train extremely deep plain ConvNets. With a mean-field-theory-based scheme, 10,000-layer networks were trained over 99\% accuracy on MNIST and 82\% on CIFAR-10. Though the models were not practical (even LeNet-5 \cite{lecun1998gradient} can reach 99.3\% accuracy on MNIST and VGG-16 can reach above 93\% on CIFAR-10), the theoretical contributions were insightful. A recent work \cite{oyedotun2020going} combined several techniques including Leaky ReLU, max-norm and careful initialization. On ImageNet, it showed that a plain ConvNet with 147M parameters could reach 74.6\% top-1 accuracy, which was 2\% lower than its reported baseline (ResNet-101, 76.6\%, 45M parameters). 

Notably, this paper is not merely a demonstration that plain models can converge reasonably well, and does not intend to train extremely deep ConvNets like ResNets. Rather, we aim to build a simple model with reasonable depth and favorable accuracy-speed trade-off, which can be simply implemented with the most common components (\eg, regular conv and BN) and simple algebra. 

\subsection{Model Re-parameterization}

DiracNet~\cite{zagoruyko2017diracnets} is a re-parameterization method related to ours. It builds deep plain models by encoding the kernel of a conv layer as $\hat{\mathrm{W}}=\text{diag}(\mathbf{a})\mathrm{I} + \text{diag}(\mathbf{b})\mathrm{W}_{\text{norm}}$, where $\hat{\mathrm{W}}$ is the eventual weight used for convolution (a 4th-order tensor viewed as a matrix), $\mathbf{a}$ and $\mathbf{b}$ are learned vectors, and $\mathrm{W}_{\text{norm}}$ is the normalized learnable kernel. Compared to ResNets with comparable amount of parameters, the top-1 accuracy of DiracNet is 2.29\% lower on CIFAR-100 (78.46\% \vs 80.75\%) and 0.62\% lower on ImageNet (72.21\% of DiracNet-34 \vs 72.83\% of ResNet-34). DiracNet differs from our method in two aspects. \textbf{1)} The training-time behavior of RepVGG is implemented by the actual dataflow through a concrete structure which can be later converted into another, while DiracNet merely uses another mathematical expression of conv kernels for easier optimization. In other words, a training-time RepVGG is a real multi-branch model, but a DiracNet is not. \textbf{2)} The performance of a DiracNet is higher than a normally parameterized plain model but lower than a comparable ResNet, while RepVGG models outperform ResNets by a large margin. Asym Conv Block (ACB) \cite{ding2019acnet}, DO-Conv \cite{cao2020conv} and ExpandNet \cite{guo2020expandnets} can also be viewed as structural re-parameterization in the sense that they convert a block into a conv. Compared to our method, the difference is that they are designed for component-level improvements and used as a drop-in replacement for conv layers in any architecture, while our structural re-parameterization is critical for training plain ConvNets, as shown in Sect. \ref{sect-1}.

\subsection{Winograd Convolution}\label{sect-wino}

RepVGG uses only $3\times3$ conv because it is highly optimized by some modern computing libraries like NVIDIA cuDNN \cite{chetlur2014cudnn} and Intel MKL \cite{intel-mkl} on GPU and CPU. Table. \ref{table-speed-kernelsize} shows the theoretical FLOPs, actual running time and computational density (measured in Tera FLoating-point Operations Per Second, TFLOPS) \footnote{As a common practice, we count a multiply-add as a single operation when counting the theoretical FLOPs, but hardware vendors like NVIDIA usually count it as two operations when reporting the TFLOPS.} tested with cuDNN 7.5.0 on a 1080Ti GPU. The theoretical computational density of $3\times3$ conv is around $4\times$ as the others, suggesting the total theoretical FLOPs is not a comparable proxy for the actual speed among different architectures. Winograd \cite{winograd} is a classic algorithm for accelerating $3\times3$ conv (only if the stride is 1), which has been well supported (and enabled by default) by libraries like cuDNN and MKL. For example, with the standard $F(2\times2, 3\times3)$ Winograd, the amount of multiplications (MULs) of a $3\times3$ conv is reduced to $\frac{4}{9}$ of the original. Since the multiplications are much more time-consuming than additions, we count the MULs to measure the computational costs with Winograd support (denoted by Wino MULs in Table. \ref{table-weak}, \ref{table-strong}). Note that the specific computing library and hardware determine whether to use Winograd for each operator because small-scale convolutions may not be accelerated due to the memory overhead. \footnote{Our results are manually tested operator-by-operator with cuDNN 7.5.0, 1080Ti. For each stride-1 $3\times3$ conv, we test its time usage along with a stride-2 counterpart of the same FLOPs. We assume the former uses $F(2\times2, 3\times3)$ Winograd if the latter runs significantly slower. Such a testing method is approximate hence the results are for reference only.}

\section{Building RepVGG via Structural Re-param}

\subsection{Simple is Fast, Memory-economical, Flexible}

There are at least three reasons for using simple ConvNets: they are fast, memory-economical and Flexible.

\textbf{Fast} \quad Many recent multi-branch architectures have lower theoretical FLOPs than VGG but may not run faster. For example, VGG-16 has $8.4\times$ FLOPs as EfficientNet-B3 \cite{efficientnet} but runs $1.8\times$ faster on 1080Ti (Table. \ref{table-weak}), which means the computational density of the former is $15\times$ as the latter. Except for the acceleration brought by Winograd conv, the discrepancy between FLOPs and speed can be attributed to two important factors that have considerable affection on speed but are not taken into account by FLOPs: the memory access cost (MAC) and degree of parallelism~\cite{ma2018shufflenet}. For example, though the required computations of branch addition or concatenation are negligible, the MAC is significant. Moreover, MAC constitutes a large portion of time usage in groupwise convolution. On the other hand, a model with high degree of parallelism could be much faster than another one with low degree of parallelism, under the same FLOPs. As multi-branch topology is widely adopted in Inception and auto-generated architectures, multiple small operators are used instead of a few large ones. A prior work \cite{ma2018shufflenet} reported that the number of fragmented operators (\ie the number of individual conv or pooling operations in one building block) in NASNET-A \cite{zoph2016neural} is 13, which is unfriendly to devices with strong parallel computing powers like GPU and introduces extra overheads such as kernel launching and synchronization. In contrast, this number is 2 or 3 in ResNets, and we make it 1: a single conv.

\begin{table}
	\caption{Speed test with varying kernel size and batch size = 32, input channels = output channels = 2048, resolution = 56$\times$56, stride = 1 on NVIDIA 1080Ti. The results of time usage are average of 10 runs after warming up the hardware.}
	\label{table-speed-kernelsize}
	\vspace{-0.2in}
	\begin{center}
		\small
			\begin{tabular}{lcccc}
				\hline
				\makecell{Kernel \\ size} 	&\makecell{Theoretical \\ FLOPs (B)}	&\makecell{Time\\usage (ms)}	&\makecell{Theoretical \\ TFLOPS}	 			\\
				\hline
				$1\times1$		&	420.9	&	84.5	&9.96\\
				$3\times3$		&	3788.1 	&	198.8	&\textbf{38.10}\\
				$5\times5$		&	10522.6	&	2092.5	&10.57\\
				$7\times7$		&	20624.4	&	4394.3	&9.38\\				
				\hline
			\end{tabular}

	\end{center}
	\vspace{-0.2in}
\end{table}

\begin{figure}
	\begin{center}
		\includegraphics[width=\linewidth]{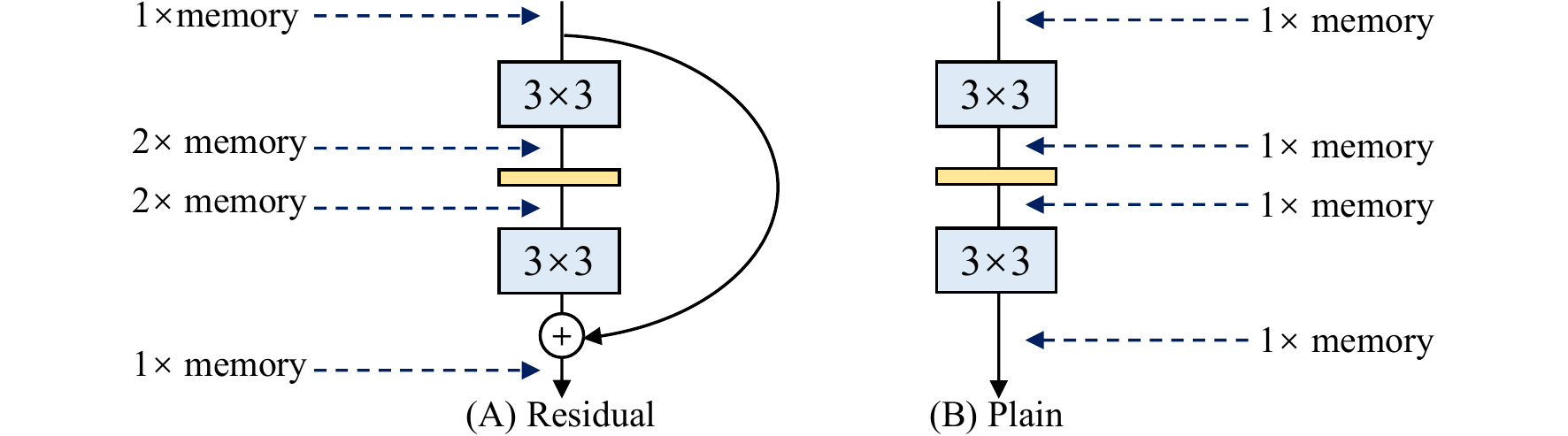}
		\vspace{-0.25in}
		\caption{Peak memory occupation in residual and plain model. If the residual block maintains the size of feature map, the peak value of extra memory occupied by feature maps will be $2\times$ as the input. The memory occupied by the parameters is small compared to the features hence ignored.}
		\label{fig-memory}
	\end{center}
	\vspace{-0.2in}
\end{figure}

\textbf{Memory-economical} \quad The multi-branch topology is memory-inefficient because the results of every branch need to be kept until the addition or concatenation, significantly raising the peak value of memory occupation. Fig.~\ref{fig-memory} shows that the input to a residual block need to be kept until the addition. Assuming the block maintains the feature map size, the peak value of extra memory occupation is 2$\times$ as the input. In contrast, a plain topology allows the memory occupied by the inputs to a specific layer to be immediately released when the operation is finished. When designing specialized hardware, a plain ConvNet allows deep memory optimizations and reduces the costs of memory units so that we can integrate more computing units onto the chip.

\textbf{Flexible} \quad The multi-branch topology imposes constraints on the architectural specification. For example, ResNet requires the conv layers to be organized as residual blocks, which limits the flexibility because the last conv layers of every residual block have to produce tensors of the same shape, or the shortcut addition will not make sense. Even worse, multi-branch topology limits the application of channel pruning~\cite{li2016pruning,he2017channel}, which is a practical technique to remove some unimportant channels, and some methods can optimize the model structure by automatically discovering the appropriate width of each layer \cite{ding2019approximated}. However, multi-branch models make pruning tricky and result in significant performance degradation or low acceleration ratio \cite{ding2019centripetal,li2016pruning,ding2018auto}. In contrast, a plain architecture allows us to freely configure every conv layer according to our requirements and prune to obtain a better performance-efficiency trade-off.

\subsection{Training-time Multi-branch Architecture}

Plain ConvNets have many strengths but one fatal weakness: the poor performance. For example, with modern components like BN \cite{ioffe2015batch}, a VGG-16 can reach over 72\% top-1 accuracy on ImageNet, which seems outdated. Our structural re-parameterization method is inspired by ResNet, which explicitly constructs a shortcut branch to model the information flow as $y = x + f(x)$ and uses a residual block to learn $f$. When the dimensions of $x$ and $f(x)$ do not match, it becomes $y = g(x) + f(x)$, where $g(x)$ is a convolutional shortcut implemented by a $1\times1$ conv. An explanation for the success of ResNets is that such a multi-branch architecture makes the model an implicit ensemble of numerous shallower models~\cite{veit2016residual}. Specifically, with $n$ blocks, the model can be interpreted as an ensemble of $2^{n}$ models, since every block branches the flow into two paths.

Since the multi-branch topology has \textit{drawbacks for inference} but the branches seem \textit{beneficial to training} \cite{veit2016residual}, we use multiple branches to make an \textit{only-training-time} ensemble of numerous models. To make most of the members shallower or simpler, we use ResNet-like identity (only if the dimensions match) and $1\times1$ branches so that the training-time information flow of a building block is $y = x + g(x) + f(x)$. We simply stack several such blocks to construct the training-time model. From the same perspective as \cite{veit2016residual}, the model becomes an ensemble of $3^{n}$ members with $n$ such blocks.

\subsection{Re-param for Plain Inference-time Model}

\begin{figure}
	\begin{center}
		\includegraphics[width=\linewidth]{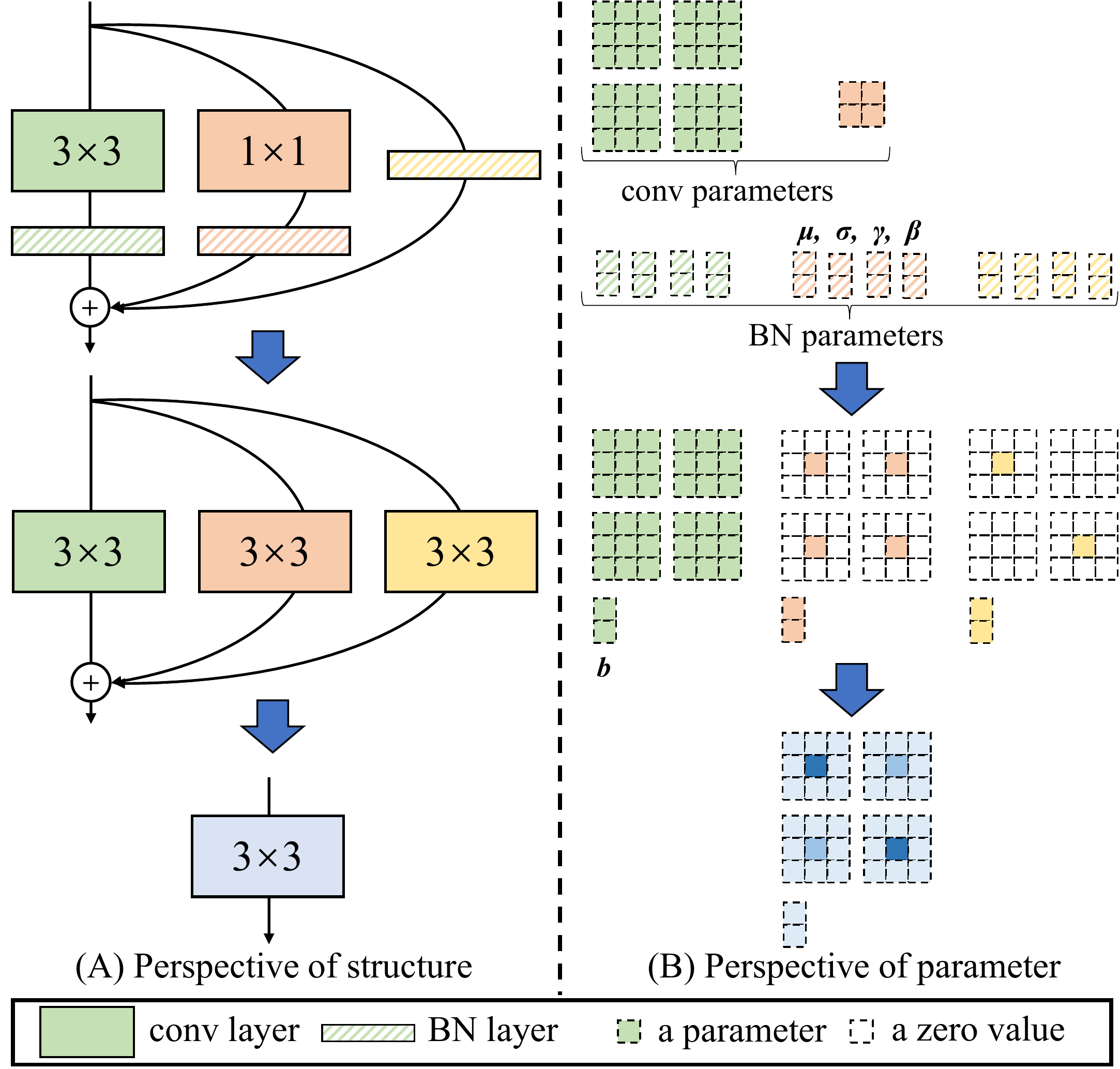}
		\vspace{-0.25in}
		\caption{Structural re-parameterization of a RepVGG block. For the ease of visualization, we assume $C_2=C_1=2$, thus the $3\times3$ layer has four $3\times3$ matrices and the kernel of $1\times1$ layer is a $2\times2$ matrix.}
		\label{fig-transform}
	\end{center}
	\vspace{-0.25in}
\end{figure}

In this subsection, we describe how to convert a trained block into a single $3\times3$ conv layer for inference. Note that we use BN in each branch before the addition (Fig.~\ref{fig-transform}). Formally, we use $\mathrm{W}^{(3)}\in\mathbb{R}^{C_2\times C_1\times 3\times 3}$ to denote the kernel of a $3\times3$ conv layer with $C_1$ input channels and $C_2$ output channels, and $\mathrm{W}^{(1)}\in\mathbb{R}^{C_2\times C_1}$ for the kernel of $1\times1$ branch. We use $\vect{\mu}^{(3)},\vect{\sigma}^{(3)},\vect{\gamma}^{(3)},\vect{\beta}^{(3)}$ as the accumulated mean, standard deviation and learned scaling factor and bias of the BN layer following $3\times3$ conv, $\vect{\mu}^{(1)},\vect{\sigma}^{(1)},\vect{\gamma}^{(1)},\vect{\beta}^{(1)}$ for the BN following $1\times1$ conv, and $\vect{\mu}^{(0)},\vect{\sigma}^{(0)},\vect{\gamma}^{(0)},\vect{\beta}^{(0)}$ for the identity branch. Let $\mathrm{M}^{(\text{1})}\in\mathbb{R}^{N\times C_1\times H_1\times W_1}$, $\mathrm{M}^{(\text{2})}\in\mathbb{R}^{N\times C_2\times H_2\times W_2}$ be the input and output, respectively, and $\ast$ be the convolution operator. If $C_1=C_2, H_1=H_2, W_1=W_2$, we have
\begin{equation}
\begin{aligned}
\mathrm{M}^{(\text{2})} &= \text{bn}(\mathrm{M}^{(\text{1})} \ast \mathrm{W}^{(3)},\vect{\mu}^{(3)},\vect{\sigma}^{(3)},\vect{\gamma}^{(3)},\vect{\beta}^{(3)}) \\
&+\text{bn}(\mathrm{M}^{(\text{1})} \ast \mathrm{W}^{(1)},\vect{\mu}^{(1)},\vect{\sigma}^{(1)},\vect{\gamma}^{(1)},\vect{\beta}^{(1)}) \\
&+\text{bn}(\mathrm{M}^{(\text{1})},\vect{\mu}^{(0)},\vect{\sigma}^{(0)},\vect{\gamma}^{(0)},\vect{\beta}^{(0)}) \,.\\
\end{aligned}
\end{equation}
Otherwise, we simply use no identity branch, hence the above equation only has the first two terms. Here $\text{bn}$ is the inference-time BN function, formally, $\forall 1\leq i \leq C_2$,
\begin{equation}
\begin{aligned}
\text{bn}(\mathrm{M},\mathbf{\vect{\mu}},\vect{\sigma},\vect{\gamma},\vect{\beta})_{:,i,:,:} = (\mathrm{M}_{:,i,:,:} - \vect{\mu}_i)\frac{\vect{\gamma}_i}{\vect{\sigma}_i} + \vect{\beta}_i \,. \\
\end{aligned}
\end{equation}

We first convert every BN and its preceding conv layer into a conv with a bias vector. Let $\{\mathrm{W}^\prime,\mathbf{b}^\prime\}$ be the kernel and bias converted from $\{\mathrm{W},\vect{\mu},\vect{\sigma},\vect{\gamma},\vect{\beta}\}$, we have
\begin{equation}\label{eq-fuse-bn}
\mathrm{W}^\prime_{i,:,:,:} = \frac{\vect{\gamma}_i}{\vect{\sigma}_i}\mathrm{W}_{i,:,:,:} \,,\quad \mathbf{b}^\prime_i = -\frac{\vect{\mu}_i \vect{\gamma}_i}{\vect{\sigma}_i} + \vect{\beta}_i \,.
\end{equation}

Then it is easy to verify that $\forall 1\leq i \leq C_2$,
\begin{equation}
\text{bn}(\mathrm{M}\ast\mathrm{W},\vect{\mu},\vect{\sigma},\vect{\gamma},\vect{\beta})_{:,i,:,:} = (\mathrm{M} \ast \mathrm{W}^\prime)_{:,i,:,:} + \mathbf{b}^\prime_i \,.
\end{equation}

This transformation also applies to the identity branch because an identity can be viewed as a $1\times1$ conv with an identity matrix as the kernel. After such transformations, we will have one $3\times3$ kernel, two $1\times1$ kernels, and three bias vectors. Then we obtain the final bias by adding up the three bias vectors, and the final $3\times3$ kernel by adding the $1\times1$ kernels onto the central point of $3\times3$ kernel, which can be easily implemented by first zero-padding the two $1\times1$ kernels to $3\times3$ and adding the three kernels up, as shown in Fig. \ref{fig-transform}. Note that the equivalence of such transformations requires the $3\times3$ and $1\times1$ layer to have the same stride, and the padding configuration of the latter shall be one pixel less than the former. For example, for a $3\times3$ layer that pads the input by one pixel, which is the most common case, the $1\times1$ layer should have padding = 0. 

\subsection{Architectural Specification}

Table. \ref{table-arch-detail} shows the specification of RepVGG including the depth and width. RepVGG is VGG-style in the sense that it adopts a plain topology and heavily uses $3\times3$ conv, but it does not use max pooling like VGG because we desire the body to have only one type of operator. We arrange the $3\times3$ layers into 5 stages, and the first layer of a stage down-samples with the stride = 2. For image classification, we use global average pooling followed by a fully-connected layer as the head. For other tasks, the task-specific heads can be used on the features produced by any layer.

We decide the numbers of layers of each stage following three simple guidelines. \textbf{1)} The first stage operates with large resolution, which is time-consuming, so we use only one layer for lower latency. \textbf{2)} The last stage shall have more channels, so we use only one layer to save the parameters. \textbf{3)} We put the most layers into the second last stage (with $14\times14$ output resolution on ImageNet), following ResNet and its recent variants \cite{he2016deep,regnet,xie2017aggregated} (\eg, ResNet-101 uses 69 layers in its $14\times14$-resolution stage). We let the five stages have 1, 2, 4, 14, 1 layers respectively to construct an instance named RepVGG-A. We also build a deeper RepVGG-B, which has 2 more layers in stage2, 3 and 4. We use RepVGG-A to compete against other lightweight and middleweight models including ResNet-18/34/50, and RepVGG-B against the high-performance ones. 

We determine the layer width by uniformly scaling the classic width setting of $[64, 128, 256, 512]$ (\eg, VGG and ResNets). We use multiplier $a$ to scale the first four stages and $b$ for the last stage, and usually set $b>a$ because we desire the last layer to have richer features for the classification or other down-stream tasks. Since RepVGG has only one layer in the last stage, a larger $b$ does not significantly increase the latency nor the amount of parameters. Specifically, the width of stage2, 3, 4, 5 is $[64a, 128a, 256a, 512b]$, respectively. To avoid large-scale conv on high-resolution feature maps, we scale down stage1 if $a<1$ but do not scale it up, so that the width of stage1 is $\text{min}(64, 64a)$.

To further reduce the parameters and computations, we may optionally interleave groupwise $3\times3$ conv layers with dense ones to trade accuracy for efficiency. Specifically, we set the number of groups $g$ for the 3rd, 5th, 7th, ..., 21st layer of RepVGG-A and the additional 23rd, 25th and 27th layers of RepVGG-B. For the simplicity, we set $g$ as 1, 2, or 4 globally for such layers without layer-wise tuning. We do not use adjacent groupwise conv layers because that would disable the inter-channel information exchange and bring a side effect \cite{zhang2018shufflenet}: outputs from a certain channel would be derived from only a small fraction of input channels. Note that the $1\times1$ branch shall have the same $g$ as the $3\times3$ conv.

\setlength{\tabcolsep}{4pt}
\begin{table}
	\caption{Architectural specification of RepVGG. Here $2\times64a$ means stage2 has 2 layers each with $64a$ channels.}
	\label{table-arch-detail}
	\vspace{-0.2in}
	\begin{center}
		\small
		\begin{tabular}{c|c|l|l}
			\hline
			Stage			&	Output size			&	 RepVGG-A 		&	RepVGG-B		\\
			\hline
			1			&	$112\times112$		&	$1\times\text{min}(64,64a)$			&	$1\times\text{min}(64,64a)$		\\
			2			&	$56\times56$		&	$2\times64a $		&	$4\times64a$		\\
			3			&	$28\times28$		&	$4\times128a$		&	$6\times128a$		\\
			4			&	$14\times14$		&	$14\times256a$		&	$16\times256a$		\\
			5			&	$7\times7$			&	$1\times512b$		&	$1\times512b$		\\
			\hline
		\end{tabular}
	\end{center}
	\vspace{-0.3in}
\end{table}
\setlength{\tabcolsep}{1.4pt}

\section{Experiments}

We compare RepVGG with the baselines on ImageNet, justify the significance of structural re-parameterization by a series of ablation studies and comparisons, and verify the generalization performance on semantic segmentation \cite{pspnet}.

\subsection{RepVGG for ImageNet Classification}\label{sect-0}

We compare RepVGG with the classic and state-of-the-art models including VGG-16 \cite{simonyan2014very}, ResNet \cite{he2016deep}, ResNeXt \cite{xie2017aggregated}, EfficientNet \cite{efficientnet}, and RegNet \cite{regnet} on ImageNet-1K \cite{deng2009imagenet}, which comprises 1.28M images for training and 50K for validation. We use EfficientNet-B0/B3 and RegNet-3.2GF/12GF as the representatives for middleweight and heavyweight state-of-the-art models, respectively. We vary the multipliers $a$ and $b$ to generate a series of RepVGG models to compare against the baselines (Table. \ref{table-model-summary}).

We first compare RepVGG against ResNets \cite{he2016deep}, which are the most common benchmarks. We use RepVGG-A0/A1/A2 for the comparisons with ResNet-18/34/50, respectively. To compare against the larger models, we construct the deeper RepVGG-B0/B1/B2/B3 with increasing width. For those RepVGG models with interleaved groupwise layers, we postfix g2/g4 to the model name. 

\setlength{\tabcolsep}{4pt}
\begin{table}
	\caption{RepVGG models defined by multipliers $a$ and $b$.}
	\label{table-model-summary}
	\vspace{-0.2in}
	\begin{center}
		\small
		\begin{tabular}{c|c|c|c}
			\hline
			Name			&	Layers of each stage	&	 $a$ 		&	$b$		\\
			\hline
			RepVGG-A0		&	1, 2, 4, 14, 1			&	0.75			&	2.5		\\
			RepVGG-A1		&	1, 2, 4, 14, 1			&	1			&	2.5		\\
			RepVGG-A2		&	1, 2, 4, 14, 1			&	1.5			&	2.75		\\
			\hline
			RepVGG-B0		&	1, 4, 6, 16, 1			&	1			&	2.5		\\
			RepVGG-B1		&	1, 4, 6, 16, 1			&	2			&	4		\\
			RepVGG-B2		&	1, 4, 6, 16, 1			&	2.5			&	5		\\
			RepVGG-B3		&	1, 4, 6, 16, 1			&	3			&	5	\\
			\hline
		\end{tabular}
	\end{center}
	\vspace{-0.3in}
\end{table}
\setlength{\tabcolsep}{1.4pt}

For training the lightweight and middleweight models, we only use the simple data augmentation pipeline including random cropping and left-right flipping, following the official PyTorch example \cite{pytorch-example}. We use a global batch size of 256 on 8 GPUs, a learning rate initialized as 0.1 and cosine annealing for 120 epochs, standard SGD with momentum coefficient of 0.9 and weight decay of $10^{-4}$ on the kernels of conv and fully-connected layers. For the heavyweight models including RegNetX-12GF, EfficientNet-B3 and RepVGG-B3, we use 5-epoch warmup, cosine learning rate annealing for 200 epochs, label smoothing \cite{szegedy2016rethinking} and mixup \cite{zhang2017mixup} (following \cite{he2019bag}), and a data augmentation pipeline of Autoaugment \cite{cubuk2019autoaugment}, random cropping and flipping. RepVGG-B2 and its g2/g4 variants are trained in both settings. We test the speed of every model with a batch size of 128 on a 1080Ti GPU \footnote{We use such a batch size because it is large enough to realize 100\% GPU utilization of every tested model to simulate the actual application scenario pursuing the maximum QPS (Queries Per Second), and our GPU memory is insufficient for EfficientNet-B3 with a batch size of 256.} by first feeding 50 batches to warm the hardware up, then 50 batches with time usage recorded. For the fair comparison, we test all the models on the same GPU, and all the conv-BN sequences of the baselines are also converted into a conv with bias (Eq. \ref{eq-fuse-bn}). 

Table. \ref{table-weak} shows the favorable accuracy-speed trade-off of RepVGG: RepVGG-A0 is 1.25\% and 33\% better than ResNet-18 in terms of accuracy and speed, RepVGG-A1 is 0.29\%/64\% better than ResNet-34, RepVGG-A2 is 0.17\%/83\% better than ResNet-50. With interleaved groupwise layers (g2/g4), the RepVGG models are further accelerated with reasonable accuracy decrease: RepVGG-B1g4 is 0.37\%/101\% better than ResNet-101, and RepVGG-B1g2 is impressively 2.66$\times$ as fast as ResNet-152 with the same accuracy. Though the number of parameters is not our primary concern, all the above RepVGG models are more parameter-efficient than ResNets. Compared to the classic VGG-16, RepVGG-B2 has only 58\% parameters, runs 10\% faster and shows 6.57\% higher accuracy. Compared to the highest-accuracy (74.5\%) VGG to the best of our knowledge trained with RePr \cite{repr} (a pruning-based training method), RepVGG-B2 outperforms by 4.28\% in accuracy.

Compared with the state-of-the-art baselines, RepVGG also shows favorable performance, considering its simplicity: RepVGG-A2 is 1.37\%/59\% better than EfficientNet-B0, RepVGG-B1 performs 0.39\% better than RegNetX-3.2GF and runs slightly faster. Notably, RepVGG models reach above 80\% accuracy with 200 epochs (Table. \ref{table-strong}), which is the first time for plain models to catch up with the state-of-the-arts, to the best of our knowledge. Compared to RegNetX-12GF, RepVGG-B3 runs 31\% faster, which is impressive considering that RepVGG does not require a lot of manpower to refine the design space like RegNet~\cite{regnet}, and the architectural hyper-parameters are set casually.

As two proxies of computational complexity, we count the theoretical FLOPs and Wino MULs as described in Sect. \ref{sect-wino}. For example, we found out that none of the conv in EfficientNet-B0/B3 is accelerated by Winograd algorithm. Table. \ref{table-weak} shows Wino MULs is a better proxy on GPU, \eg, ResNet-152 runs slower than VGG-16 with lower theoretical FLOPs but higher Wino MULs. Of course, the actual speed should always be the golden standard.

\setlength{\tabcolsep}{4pt}
\begin{table}
	\caption{Results trained on ImageNet with simple data augmentation in 120 epochs. The speed is tested on 1080Ti with a batch size of 128, full precision (fp32), and measured in examples/second. We count the theoretical FLOPs and Wino MULs as described in Sect. \ref{sect-wino}. The baselines are our implementations with the same training settings.}
	\label{table-weak}
	\vspace{-0.2in}
	\begin{center}		
		\small
		\begin{tabular}{lccccccc}
			\hline
			Model			&	\makecell{Top-1 \\ acc}		&	 \makecell{Speed} 	&	\makecell{Params \\ (M)}	&\makecell{Theo \\ FLOPs \\ (B)}		&\makecell{Wino \\ MULs \\ (B)}	\\
			\hline
			\textbf{RepVGG-A0} & 72.41 & 3256 & 8.30 & 1.4 & 0.7 \\
			ResNet-18 & 71.16 & 2442 & 11.68 & 1.8 & 1.0 \\
			\hline
			\textbf{RepVGG-A1} & 74.46 & 2339 & 12.78 & 2.4 & 1.3 \\
			\textbf{RepVGG-B0} & 75.14 & 1817 & 14.33 & 3.1 & 1.6 \\
			ResNet-34 & 74.17 & 1419 & 21.78 & 3.7 & 1.8 \\
			\hline
			\textbf{RepVGG-A2} & 76.48 & 1322 & 25.49 & 5.1 & 2.7 \\
			\textbf{RepVGG-B1g4} & 77.58 & 868 & 36.12 & 7.3 & 3.9 \\
			EfficientNet-B0 & 75.11 & 829 & 5.26 & 0.4 & - \\
			\hline
			\textbf{RepVGG-B1g2} & 77.78 & 792 & 41.36 & 8.8 & 4.6 \\
			ResNet-50 & 76.31 & 719 & 25.53 & 3.9 & 2.8 \\
			\hline 
			\textbf{RepVGG-B1} & 78.37 & 685 & 51.82 & 11.8 & 5.9 \\
			RegNetX-3.2GF & 77.98 & 671 & 15.26 & 3.2 & 2.9 \\
			\hline
			\textbf{RepVGG-B2g4} & 78.50 & 581 & 55.77 & 11.3 & 6.0 \\
			ResNeXt-50 & 77.46 & 484 & 24.99 & 4.2 & 4.1 \\
			\hline
			\textbf{RepVGG-B2} & 78.78 & 460 & 80.31 & 18.4 & 9.1 \\
			ResNet-101 & 77.21 & 430 & 44.49 & 7.6 & 5.5 \\
			VGG-16 & 72.21 & 415 & 138.35 & 15.5 & 6.9 \\
			ResNet-152 & 77.78 & 297 & 60.11 & 11.3 & 8.1 \\
			ResNeXt-101 & 78.42 & 295 & 44.10 & 8.0 & 7.9 \\
			\hline
		\end{tabular}
	\end{center}
	\vspace{-0.2in}
\end{table}
\setlength{\tabcolsep}{1.4pt}

\setlength{\tabcolsep}{4pt}
\begin{table}
	\caption{Results on ImageNet trained in 200 epochs with Autoaugment \cite{cubuk2019autoaugment}, label smoothing and mixup.}
	\label{table-strong}
	\vspace{-0.2in}
	\begin{center}
		\small
		\begin{tabular}{lccccccc}
			\hline
			Model			&	Acc		&	 \makecell{Speed} 	&	\makecell{Params}	&\makecell{FLOPs}		&\makecell{MULs}	\\
			\hline
			\textbf{RepVGG-B2g4} & 79.38 & 581 & 55.77 & 11.3 & 6.0 \\
			\textbf{RepVGG-B3g4} & 80.21 & 464 & 75.62 & 16.1 & 8.4 \\
			\textbf{RepVGG-B3} & 80.52 & 363 & 110.96 & 26.2 & 12.9 \\
			RegNetX-12GF & 80.55 & 277 & 46.05 & 12.1 & 10.9 \\
			EfficientNet-B3 & 79.31 & 224 & 12.19 & 1.8 & - \\
			\hline
		\end{tabular}
	\end{center}
	\vspace{-0.35in}
\end{table}
\setlength{\tabcolsep}{1.4pt}

\subsection{Structural Re-parameterization is the Key}\label{sect-1}

In this subsection, we verify the significance of our structural re-parameterization technique (Table. \ref{table-ablation}). All the models are trained from scratch for 120 epochs with the same simple training settings described above. First, we conduct ablation studies by removing the identity and/or $1\times1$ branch from every block of RepVGG-B0. With both branches removed, the training-time model degrades into an ordinary plain model and only achieves 72.39\% accuracy. The accuracy is lifted to 73.15\% with $1\times1$ or 74.79\% with identity. The accuracy of the full featured RepVGG-B0 is 75.14\%, which is 2.75\% higher than the ordinary plain model. Seen from the inference speed of the training-time (\ie, not yet converted) models, removing the identity and $1\times1$ branches via structural re-parameterization brings significant speedup.

Then we construct a series of variants and baselines for comparison on RepVGG-B0 (Table. \ref{table-variants}). Again, all the models are trained from scratch in 120 epochs.
\begin{itemize}[noitemsep,nolistsep,topsep=0pt,parsep=0pt,partopsep=0pt]
	\item \textbf{Identity w/o BN} removes the BN in identity branch.
	\item \textbf{Post-addition BN} removes the BN layers in the three branches and appends a BN layer after the addition. In other words, the position of BN is changed from pre-addition to post-addition.
	\item \textbf{+ReLU in branches} inserts ReLU into each branch (after BN and before addition). Since such a block cannot be converted into a single conv layer, it is of no practical use, and we merely desire to see whether more nonlinearity will bring higher performance.
	\item \textbf{DiracNet} \cite{zagoruyko2017diracnets} adopts a well-designed re-parameterization of conv kernels, as introduced in Sect. \ref{sect-related-work-2}. We use its official PyTorch code to build the layers to replace the original $3\times3$ conv.
	\item \textbf{Trivial Re-param} is a simpler re-parameterization of conv kernels by directly adding an identity kernel to the $3\times3$ kernel, which can be viewed a degraded version of DiracNet ($\hat{\mathrm{W}}=\mathrm{I} + \mathrm{W}$ \cite{zagoruyko2017diracnets}).
	\item \textbf{Asymmetric Conv Block} (ACB) \cite{ding2019acnet} can be viewed as another form of structural re-parameterization. We compare with ACB to see whether the improvement of our structural re-parameterization is due to the component-level over-parameterization (\ie, the extra parameters making every $3\times3$ conv stronger).
	\item \textbf{Residual Reorg} builds each stage by re-organizing it in a ResNet-like manner (2 layers per block). Specifically, the resultant model has one $3\times3$ layer in the first and last stages and 2, 3, 8 residual blocks in stage2, 3, 4, and uses shortcuts just like ResNet-18/34.
\end{itemize}

We reckon the superiority of structural re-param over DiractNet and Trivial Re-param lies in the fact that the former relies on the actual dataflow through a concrete structure with nonlinear behavior (BN), while the latter merely uses another mathematical expression of conv kernels. The former ``re-param'' means ``using the params of a structure to parameterize another structure'', but the latter means ``computing the params first with another set of params, then using them for other computations''. With nonlinear components like a training-time BN, the former cannot be approximated by the latter. As evidences, the accuracy is decreased by removing the BN and improved by adding ReLU. In other words, though a RepVGG block can be equivalently converted into a single conv for inference, the inference-time equivalence does not imply the training-time equivalence, as we cannot construct a conv layer to have the same training-time behavior as a RepVGG block. 

The comparison with ACB suggests the success of RepVGG should not be simply attributed to the effect of over-parameterization of every component, since ACB uses more parameters but yields inferior performance. As a double check, we replace every $3\times3$ conv of ResNet-50 with a RepVGG block and train from scratch for 120 epochs. The accuracy is 76.34\%, which is merely 0.03\% higher than the ResNet-50 baseline, suggesting that RepVGG-style structural re-parameterization is not a generic over-parameterization technique, but a methodology critical for training powerful plain ConvNets. Compared to Residual Reorg, a real residual network with the same number of $3\times3$ conv and additional shortcuts for both training and inference, RepVGG outperforms by 0.58\%, which is not surprising since RepVGG has far more branches. For example, the branches make stage4 of RepVGG an ensemble of $2\times3^{15}=2.8\times10^7$ models \cite{veit2016residual}, while the number for Residual Reorg is $2^8=256$.

\setlength{\tabcolsep}{4pt}
\begin{table}
	\caption{Ablation studies with 120 epochs on RepVGG-B0. The inference speed w/o re-param (examples/s) is tested with the models before conversion (batch size=128). Note again that all the models have the same final structure.}
	\label{table-ablation}
	\vspace{-0.2in}
	\begin{center}
		\small
		\begin{tabular}{lcccc}
			\hline
				\makecell{Identity \\ branch}		&	 \makecell{$1\times1$ \\ branch} 	&	Accuracy	&	\makecell{Inference speed \\ w/o re-param}\\
			\hline
								&						&	72.39			&	1810\\
				\checkmark 		&						&	74.79			&	1569\\
						 		&	\checkmark			&	73.15			&	1230\\
				\checkmark 		&	\checkmark			&	\textbf{75.14}	&	1061\\
			\hline
		\end{tabular}
	\end{center}
	\vspace{-0.2in}
\end{table}
\setlength{\tabcolsep}{1.4pt}

\setlength{\tabcolsep}{4pt}
\begin{table}
	\caption{Comparison with variants and baselines on RepVGG-B0 trained in 120 epochs.}
	\label{table-variants}
	\vspace{-0.2in}
	\begin{center}
		\small
		\begin{tabular}{lc}
			\hline
			Variant and baseline&	Accuracy		\\
			\hline
			Identity w/o BN		&	74.18			\\
			Post-addition BN				&	73.52			\\
			Full-featured reparam	&	\textbf{75.14}	\\
			+ReLU in branch 	&	75.69			\\
			\hline
			DiracNet \cite{zagoruyko2017diracnets}			&	73.97			\\
			Trivial Re-param	&	73.51			\\
			ACB \cite{ding2019acnet}	&	73.58			\\
			Residual Reorg		&	74.56	\\
			\hline
		\end{tabular}
	\end{center}
	\vspace{-0.3in}
\end{table}
\setlength{\tabcolsep}{1.4pt}

\subsection{Semantic Segmentation}
We verify the generalization performance of ImageNet-pretrained RepVGG for semantic segmentation on Cityscapes \cite{cityscapes} (Table. \ref{table-seg}). We use the PSPNet \cite{pspnet} framework, a poly learning rate policy with base of 0.01 and power of 0.9, weight decay of $10^{-4}$ and a global batch size of 16 on 8 GPUs for 40 epochs. For the fair comparison, we only change the ResNet-50/101 backbone to RepVGG-B1g2/B2 and keep other settings identical. Following the official PSPNet-50/101 \cite{pspnet} which uses dilated conv in the last two stages of ResNet-50/101, we also make all the $3\times3$ conv layers in the last two stages of RepVGG-B1g2/B2 dilated. However, the current inefficient implementation of $3\times3$ dilated conv (though the FLOPs is the same as $3\times3$ regular conv) slows down the inference. For the ease of comparison, we build another two PSPNets (denoted by \textit{fast}) with dilation only in the last 5 layers (\ie, the last 4 layers of stage4 and the only layer of stage5), so that the PSPNets run slightly faster than the ResNet-50/101-backbone counterparts. RepVGG backbones outperform ResNet-50 and ResNet-101 by 1.71\% and 1.01\% respectively in mean IoU with higher speed, and RepVGG-B1g2-fast outperforms the ResNet-101 backbone by 0.37 in mIoU and runs 62\% faster. Interestingly, dilation seems more effective for larger models, as using more dilated conv layers does not improve the performance compared to RepVGG-B1g2-fast, but raises the mIoU of RepVGG-B2 by 1.05\% with reasonable slowdown.

\setlength{\tabcolsep}{4pt}
\begin{table}
	\caption{Semantic segmentation on Cityscapes \cite{cityscapes} tested on the \textit{validation} subset. The speed (examples/second) is tested with a batch size of 16, full precision (fp32), and input resolution of 713$\times$713 on the same 1080Ti GPU.}
	\label{table-seg}
	\vspace{-0.2in}
	\begin{center}
		\small
		\begin{tabular}{lccccccc}
			\hline
			Backbone		&	Mean IoU			&	Mean pixel acc 	& Speed		\\
			\hline
			\textbf{RepVGG-B1g2-fast}	&	78.88	&	96.19	&	10.9	\\
			ResNet-50					&	77.17	&	95.99	&	10.4	\\
			\textbf{RepVGG-B1g2}		&	78.70	&	96.27	&	8.0	\\
			\hline
			\textbf{RepVGG-B2-fast}		&	79.52	&	96.36	&	6.9	\\
			ResNet-101				&	78.51	&	96.30		&	6.7	\\
			\textbf{RepVGG-B2}		&	80.57	&	96.50		&	4.5 	\\
			\hline
		\end{tabular}
	\end{center}
	\vspace{-0.3in}
\end{table}
\setlength{\tabcolsep}{1.4pt}

\subsection{Limitations}

RepVGG models are fast, simple and practical ConvNets designed for the maximum speed on GPU and specialized hardware, less concerning the number of parameters. They are more parameter-efficient than ResNets but may be less favored than the mobile-regime models like MobileNets \cite{mbv1,mbv2,mbv3} and ShuffleNets \cite{zhang2018shufflenet,ma2018shufflenet} for low-power devices.

\section{Conclusion}

We proposed RepVGG, a simple architecture with a stack of $3\times3$ conv and ReLU, which is especially suitable for GPU and specialized inference chips. With our structural re-parameterization method, it reaches over 80\% top-1 accuracy on ImageNet and shows favorable speed-accuracy trade-off compared to the state-of-the-art models.

{\small
\bibliographystyle{ieee_fullname}
\bibliography{repvggbib}
}

\end{document}